\documentclass{article}
\PassOptionsToPackage{numbers, compress}{natbib}

\usepackage[preprint]{neurips_2024}

\usepackage[utf8]{inputenc} 
\usepackage[T1]{fontenc}    
\usepackage{graphicx}
\usepackage{hyperref}       
\usepackage{url}            
\usepackage{booktabs}       
\usepackage{amsfonts}       
\usepackage{nicefrac}       
\usepackage{microtype}      
\usepackage{xcolor}         
\usepackage{enumitem}
\usepackage{amsmath}
\usepackage{wrapfig}
\usepackage{multirow}
\usepackage{authblk}
\usepackage{subcaption}

\def\rvepsilon{{\mathbf{\epsilon}}}
\def\rvtheta{{\mathbf{\theta}}}
\def\rvphi{{\mathbf{\phi}}}

\def\rve{{\mathbf{e}}}

\def\rvr{{\mathbf{r}}}

\def\rvx{{\mathbf{x}}}

\title{Toffee: Efficient Million-Scale Dataset Construction for Subject-Driven Text-to-Image Generation}

\author[1]{\textbf{Yufan Zhou}}
\author[1]{\textbf{Ruiyi Zhang}}
\author[2]{\textbf{Kaizhi Zheng}}
\author[1]{\textbf{Nanxuan Zhao}}
\author[1]{\textbf{Jiuxiang Gu}}
\author[1]{\textbf{Zichao Wang}}
\author[2]{~~~~~~~~~~\textbf{Xin Eric Wang}}
\author[1]{\textbf{Tong Sun}}

\affil[1]{Adobe Research}
\affil[2]{University of California, Santa Cruz}

\affil[1]{\texttt{\{yufzhou, ruizhang, nanxuanz, jigu, jackwa, tsun\}@adobe.com}}
\affil[2]{\texttt{\{kzheng31, xwang366\}@ucsc.edu}}

\begin{document}

\maketitle

\begin{abstract}
In subject-driven text-to-image generation, recent works have achieved superior performance by training the model on synthetic datasets containing numerous image pairs. 
Trained on these datasets, generative models can produce text-aligned images for specific subject from arbitrary testing image in a zero-shot manner. They even outperform methods which require additional fine-tuning on testing images. 
However, the cost of creating such datasets is prohibitive for most researchers. 
To generate a single training pair, current methods fine-tune a pre-trained text-to-image model on the subject image to capture fine-grained details, then use the fine-tuned model to create images for the same subject based on creative text prompts.
Consequently, constructing a large-scale dataset with millions of subjects can require hundreds of thousands of GPU hours.
To tackle this problem, we propose Toffee, an efficient method to construct datasets for subject-driven editing and generation. Specifically, our dataset construction does not need any subject-level fine-tuning. After pre-training two generative models, we are able to generate infinite number of high-quality samples. We construct the first large-scale dataset for subject-driven image editing and generation, which contains 5 million image pairs, text prompts, and masks. Our dataset is 5 times the size of previous largest dataset, yet our cost is tens of thousands of GPU hours lower. To test the proposed dataset, we also propose a model which is capable of both subject-driven image editing and generation. By simply training the model on our proposed dataset, it obtains competitive results,
illustrating the effectiveness of the proposed dataset construction framework.
\end{abstract}    
\section{Introduction}
Subject-driven text-to-image generation aims at generating creative contents for a specific concept contained in single or few user-provided images. It has attracted significant interest recently, as pre-trained text-to-image generation models~\citep{betker2023improving, chang2023muse,ding2021cogview,ramesh2022hierarchical,ramesh2021zero,rombach2022LDM, saharia2022imagen,yu2022scaling,yu2023CM3Leon,zhou2023shifted} often fails to generate images for specific subject which may only appear in single testing image. 
Various methods have been proposed for this task. 
Some methods~\citep{kumari2022multi,qiu2023oft,ruiz2023dreambooth} propose to fine-tune a pre-trained text-to-image generation model on testing images. Because the fine-grained subject details has already been captured during fine-tuning, the fine-tuned model can be used to generate creative images for the specific subject.
Some methods propose to use embeddings to represent the subject~\citep{gal2022textualinversion,gal2023e4t,li2023blipdiffusion,ma2023subjectdiffusion,wei2023elite,zhou2023profusion}. The embeddings are obtained through optimization or an image encoder, and will be injected into the text-to-image generation model in various ways to perform subject-driven text-to-image generation.

\begin{figure}[t!]
    \centering
    \includegraphics[width=0.95\textwidth]{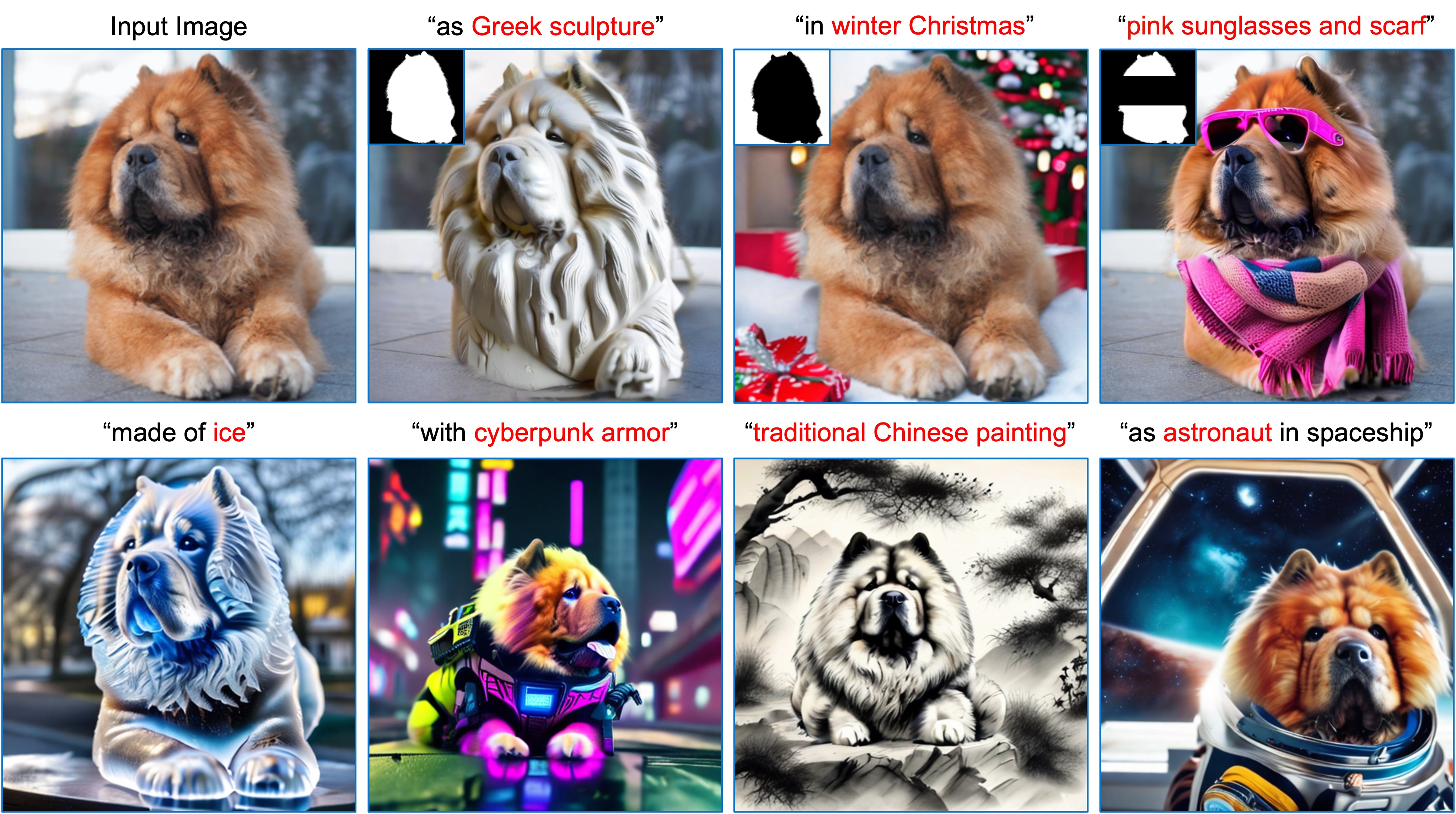}
    \vspace{-0.05in}
    \caption{Subject-driven editing and generation examples from our model, which is trained on our proposed dataset and does not require any fine-tuning at test-time. Editing masks are also presented.}
    \label{fig:gen_large}
    \vspace{-0.2in}
\end{figure}

\begin{figure}[t!]
    \centering
     \includegraphics[width=0.7\linewidth]{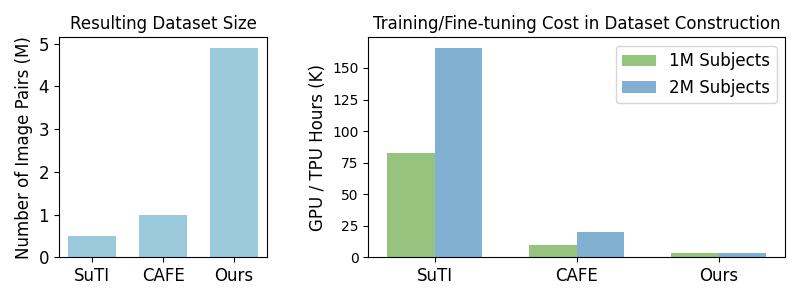}
    \vspace{-0.1in}
      \caption{Comparison of resulting dataset size and construction cost.}
      \label{fig:dataset_stat}
    \vspace{-0.25in}
\end{figure}


Different from aforementioned methods, SuTI~\citep{chen2023suti} and CAFE~\citep{zhou2023cafe} obtain impressive subject-driven generation results by training text-to-image generation models on large-scale datasets which contain paired images. 
In these datasets, each paired images depict the same subject but differ in terms of style, background, etc. By training on such datasets, the model is able to abstract high-level subject information and generalize, thus can efficiently generate images with different contexts and styles for a given testing subject, without any test-time fine-tuning.
However, one major drawback that prevents these methods from being widely used is that, although it does not require test-time fine-tuning, the dataset construction cost is actually prohibitive. 
In dataset construction stage, SuTI and CAFE require subject-level fine-tuning to generate training pairs, meaning that they need to fine-tune a text-to-image generation model on every subject and use the fine-tuned model to generate images prepared for large-scale training. Constructing large-scale dataset using methods from SuTI and CAFE can cost tens of thousands of GPU hours. Thus, they are not suitable for most researchers in the community who may not have much computational resource.

In this paper, we propose Toffee, a method \textit{TOwards eFFiciEnt datasEt construction} for subject-driven text-to-image generation. Different from existing methods~\cite{chen2023suti,zhou2023cafe} which require model fine-tuning at subject-level in dataset construction, Toffee only pre-trains two generative models. 
In other words, to construct a dataset with N subjects, previous methods~\cite{chen2023suti,zhou2023cafe} require O(N) fine-tuning steps, while Toffee requires O(1) fine-tuning steps, which is extremely important in large-scale dataset construction. 
A more straightforward comparison is provided in Figure~\ref{fig:dataset_stat}, where 
we calculate the dataset construction cost according to the details provided in ~\cite{chen2023suti, zhou2023cafe}. To construct a dataset with 1 million subjects, the fine-tuning cost for SuTI is approximately 83,000 TPU hours, while CAFE requires around 10,000 GPU hours. These computation costs scale linearly with the number of subjects. In contrast, our dataset construction pipeline requires less than 3,000 GPU hours for pre-training, with no additional costs as the number of subjects increases. Thus our efficiency advantage becomes even more pronounced as the dataset scale grows.

With the proposed method, we construct a large-scale dataset which not only contains paired images for subject-driven generation, but also contains image editing pairs and masks for subject-driven editing task. By training a unified model on the proposed dataset, we obtain competitive results on subject-driven generation without any test-time fine-tuning, illustrating the effectiveness of the proposed method. Our contributions can be summarized as follows:

\begin{itemize}[noitemsep,topsep=0pt]
    \item We propose Toffee, a novel method that leads to efficient and high-quality dataset construction for subject-driven text-to-image generation. Compared to previous methods, Toffee can save tens of thousands of GPU hours in constructing large-scale dataset for subject-driven generation.
    \item We construct Toffee-5M, the first large-scale dataset for subject-driven image generation and editing tasks. Compared to related datasets, our dataset is 5 times the size of the previous largest dataset. Our pre-trained models for the dataset construction pipeline will be made publicly available to support and advance research in related domains;
    \item We propose a new model, ToffeeNet, which is capable of both subject-driven image editing and generation with single unified model. After training the proposed model with our new dataset, we obtain competitive results in subject-driven generation within seconds, without the need of test-time fine-tuning. Extensive ablation studies are also conducted;
\end{itemize}

\section{Method}
In this section, we will present the details of our proposed Toffee. Specifically, we will first present our proposed dataset construction pipeline, and then present our new model which is capable of both image editing and generation. Training our proposed model on the new dataset enables subject-driven generation without any test-time fine-tuning, given arbitrary subject image and text during inference.

\subsection{Dataset Construction}
Although existing datasets like MVImageNet~\citep{yu2023mvimgnet} contain multi-view images for single subject, there is no color or style change between paired images, which prevents the model trained on these datasets from generating creative contents with respect to arbitrary text. Hence, our goal is to efficiently construct a large-scale dataset containing image pairs, where both images from each pair should contain same subject, while they should differ in terms of style, color, background, etc.
Training models on such dataset leads to subject-driven generation without the need of test-time fine-tuning.

\begin{figure}[t!]
    \centering
    \includegraphics[width=0.95\linewidth]{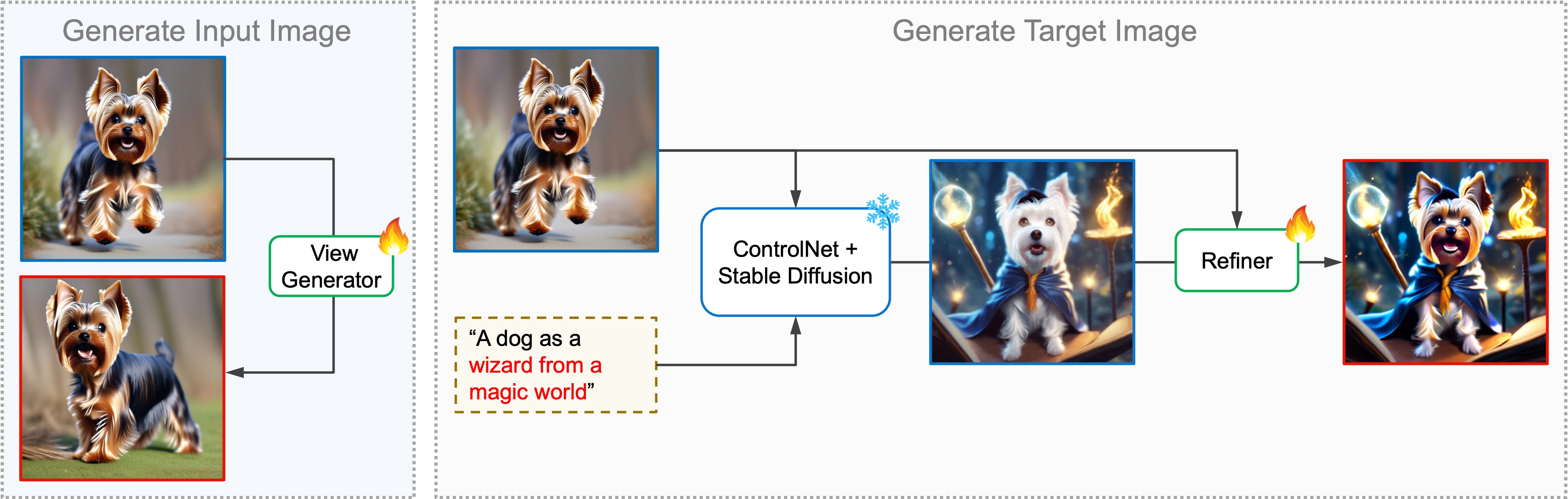}
    \vspace{-0.05in}
    \caption{The proposed dataset construction framework. 
    }
    \label{fig:dataset_construction_pipeline}
    \vspace{-0.1in}
\end{figure}

Our proposed dataset construction framework is illustrated in Figure~\ref{fig:dataset_construction_pipeline}. Given a subject image, we feed the subject image into a pre-trained diffusion model with ControlNet~\citep{zhang2023controlnet}, which generates text-aligned image without fine-grained subject details. Then, the Refiner refines the subject details in the image. Finally, the View Generator generates an image of the same subject with a different view. Both Refiner and View Generator are diffusion models trained by us, which will only be used in dataset construction. After pre-training Refiner and View Generator, data samples can be generated without any subject-level fine-tuning, which can significantly reduce computational requirements of dataset construction. For example, if we were to create the dataset using previous methods, such as DreamBooth~\citep{ruiz2023dreambooth} similar to what SuTI~\cite{chen2023suti} does, each additional one million pairs would require tens of thousands of extra GPU hours for subject-level fine-tuning.

\begin{figure}[t!]
    \centering
    \includegraphics[width=0.95\linewidth]{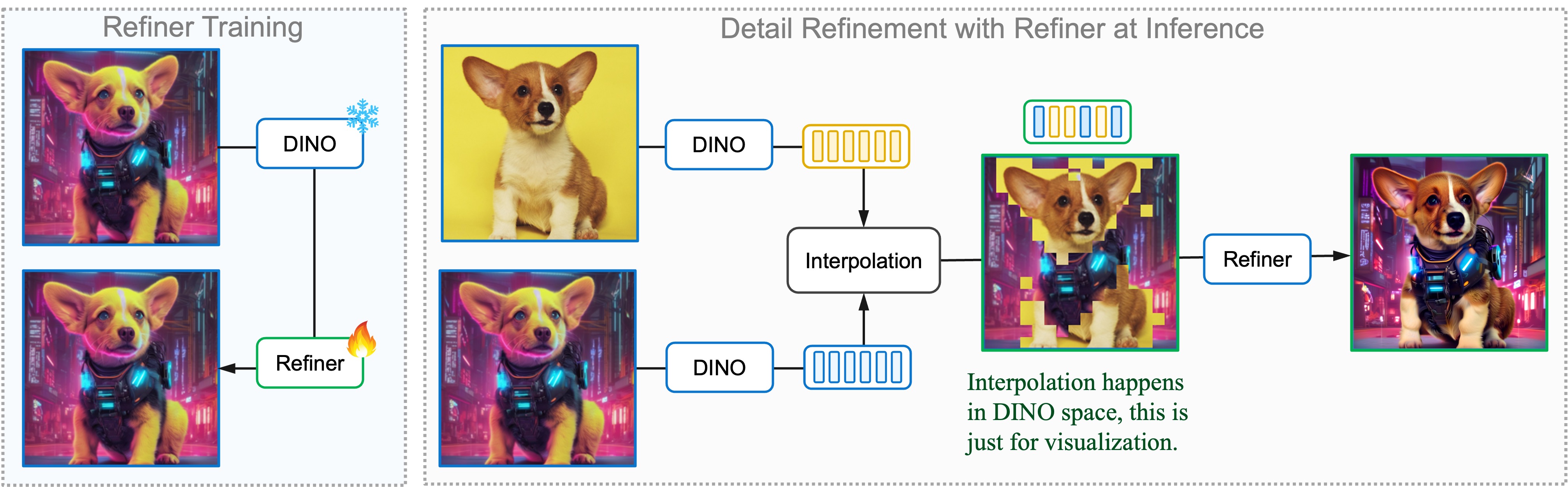}
    \vspace{-0.05in}
    \caption{Illustration of training and inference with our Refiner model.}
    \label{fig:refiner}
    \vspace{-0.2in}
\end{figure}

\paragraph{Refiner}
As shown in previous research~\cite{caron2021dino,oquab2023dinov2}, distance between patch embeddings from pre-trained DINO encoder~\citep{caron2021dino,oquab2023dinov2} can be used to perform semantic matching between image patches. Based on this interesting finding, we propose a Refiner method which can refine subject details in low-quality image pairs. Training and inference with the proposed Refiner is illustrated in Figure \ref{fig:refiner}.

Refiner is a diffusion model which is trained with the diffusion loss:
\begin{align}\label{eq:refiner_loss}
    \mathcal{L}_{\rvtheta} = \mathbb{E}\left[ \Vert \rvepsilon - R_{\theta}(f(\rvx),\rvx_t, t) \Vert^2] \right]
\end{align}
where $R_\rvtheta$ denotes Refiner, $f$ denotes pre-trained DINO image enocder, $\rvepsilon \sim \mathcal{N}(0, \mathbf{I})$ denotes randomly sampled noise,  $\rvx_t$ denotes noised image at time t, $\rvx$ denotes image without noise. Briefly speaking, our Refiner is trained to take DINO embeddings as inputs and reconstruct corresponding images. The DINO embeddings are injected into UNet through cross-attention layers.

We now present how the Refiner enhances quality of generated image pairs at inference. 
Let $\rvx, \rvx^\prime$ be subject and generated images, respectively, with corresponding DINO embeddings $f(\rvx), f(\rvx^\prime)$. Each DINO embedding is a sequence of vectors, we use $f_i(\rvx)$ to denote $i^{th}$ vector of $f(\rvx)$, which corresponds to a specific patch from image $\rvx$. 

For each patch in $\rvx^\prime$, we first find the most similar patch from $\rvx$ by patch embedding similarity:
\begin{align}
    \rve_i = \text{argmax}_{j} \text{Sim}(f_j(\rvx), f_i(\rvx^\prime))
\end{align}
where $\text{Sim}$ stands for cosine similarity.

Then we obtain a mixed DINO embedding $\rvr$ by performing linear combination between $f(\rvx^\prime)$ and $\rve$ only on highly similar patches:
\begin{align}
    \rvr_i =
    \begin{cases}
        \alpha \rve_i + (1 - \alpha) f_i(\rvx^\prime), & \text{if}\ \text{Sim}(\rve_i, f_i(\rvx^\prime))\geq \beta \\
        f_i(\rvx^\prime), & \text{otherwise}
    \end{cases}
\end{align}
where $0 \leq \alpha \leq 1$ and $-1 \leq \beta \leq 1$ are hyper-parameters. The mixed DINO embedding will be fed into the Refiner, leading to the generation of a harmonized image with refined identity. 

\begin{figure}[t!]
    \centering
    \includegraphics[width=0.95\linewidth]{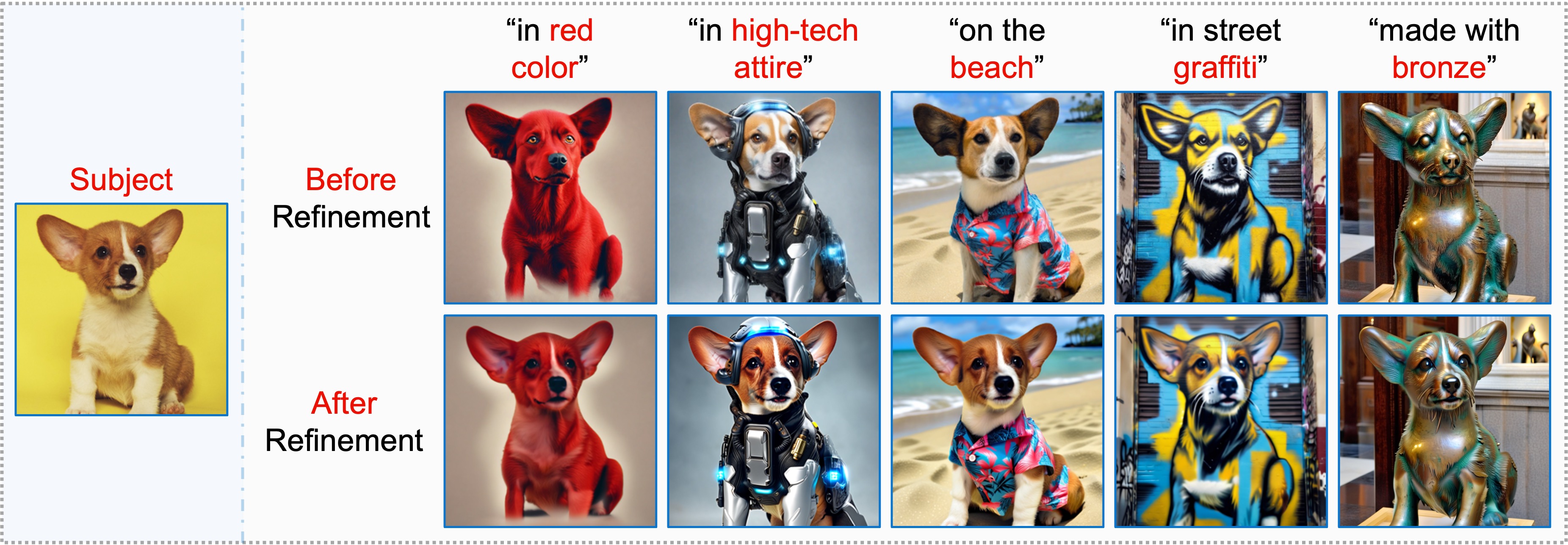}
    \vspace{-0.05in}
    \caption{Our Refiner can refine the subject details, without the loss of text-alignment.}
    \label{fig:refine_example}
    \vspace{-0.2in}
\end{figure}

As shown in Figure~\ref{fig:refiner} and Figure \ref{fig:refine_example}, corresponding patches will be successfully identified. The Refiner can improve the subject details without the loss of text-alignment. The desired differences between the target and input images in terms of style, color, texture, background, and other elements will be maintained.
Note that if we directly perform patch interpolation or replacement in pixel space, the resulting image will be of low-quality. We also find that, in practice, applying SDEdit~\citep{meng2021sdedit} improves the quality of the final image. At inference, the denoising process of our Refiner starts from noisy image $\rvx^\prime_t$ where $t<T$.

\paragraph{View Generator}
Although we can now obtain high-quality image pairs with various attribution changes, readers may notice that the subject image and target image share similar subject view and pose. To introduce more diversity into our dataset, we propose to train a View Generator.

View Generator is a diffusion model trained on multi-view image dataset. Since we only care about view change in training View Generator, dataset lacking style changes such as MVImageNet~\citep{yu2023mvimgnet} can be utilized. Specifically, let $\rvx$ be a subject image from the dataset, $\tilde{\rvx}$ be a randomly sampled image of same subject, the View Generator $G_{\rvphi}$ is trained to generate $\tilde{\rvx}$ based on DINO embedding $f(\rvx)$:
\begin{align}\label{eq:view_gen_loss}
    \mathcal{L}_{\rvphi} = \mathbb{E}\left[ \Vert \rvepsilon - G_{\theta}(f(\rvx),\tilde{\rvx}_t, t) \Vert^2] \right]
\end{align}

\paragraph{Toffee-5M Dataset}
\begin{wrapfigure}{r}{0.45\textwidth}
    \centering
    \vspace{-0.1in}
    \includegraphics[width=0.99\linewidth]{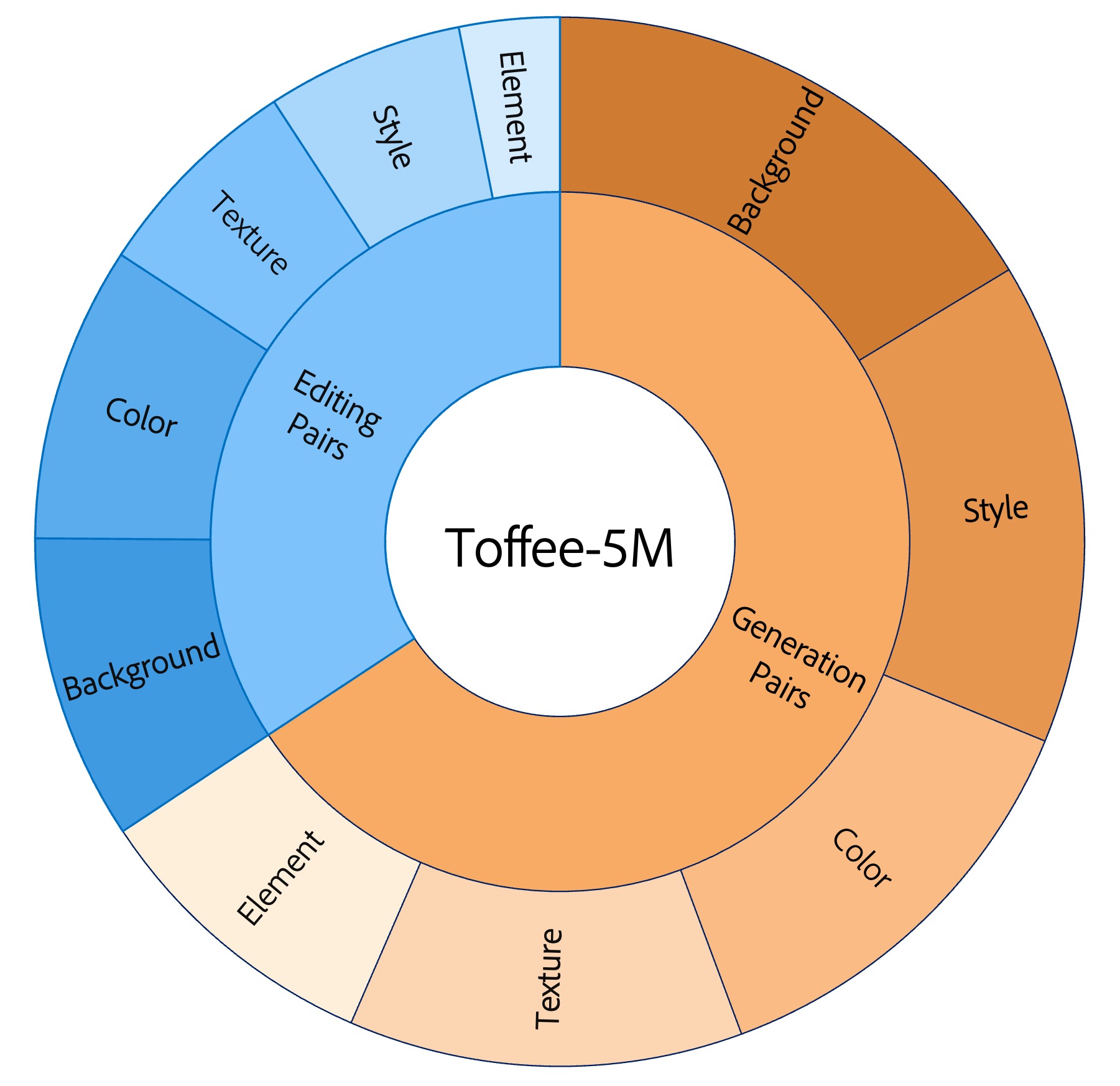}
    \caption{Taxonomy of our Toffee-5M.}
    \label{fig:dataset}
    \vspace{-0.2in}
\end{wrapfigure}

Using the trained Refiner and View Generator, we are able to efficiently obtain high-quality image pairs. 
To start with, we generate a set of subject images with pre-trained Stable Diffusion XL~\citep{podell2023sdxl,rombach2021high}, which includes 2 millions of subjects spanning around 200 different classes. Some of the classes are taken from ImageNet~\citep{deng2009imagenet} classes, while others are created by us to represent objects commonly found in daily life.
Then we collect some manually designed text prompts from workers through a platform named Upwork. We generate more prompts by prompting pre-trained Llama-2-70B model. 

\begin{figure}[t!]
    \centering
    \includegraphics[width=0.95\linewidth]{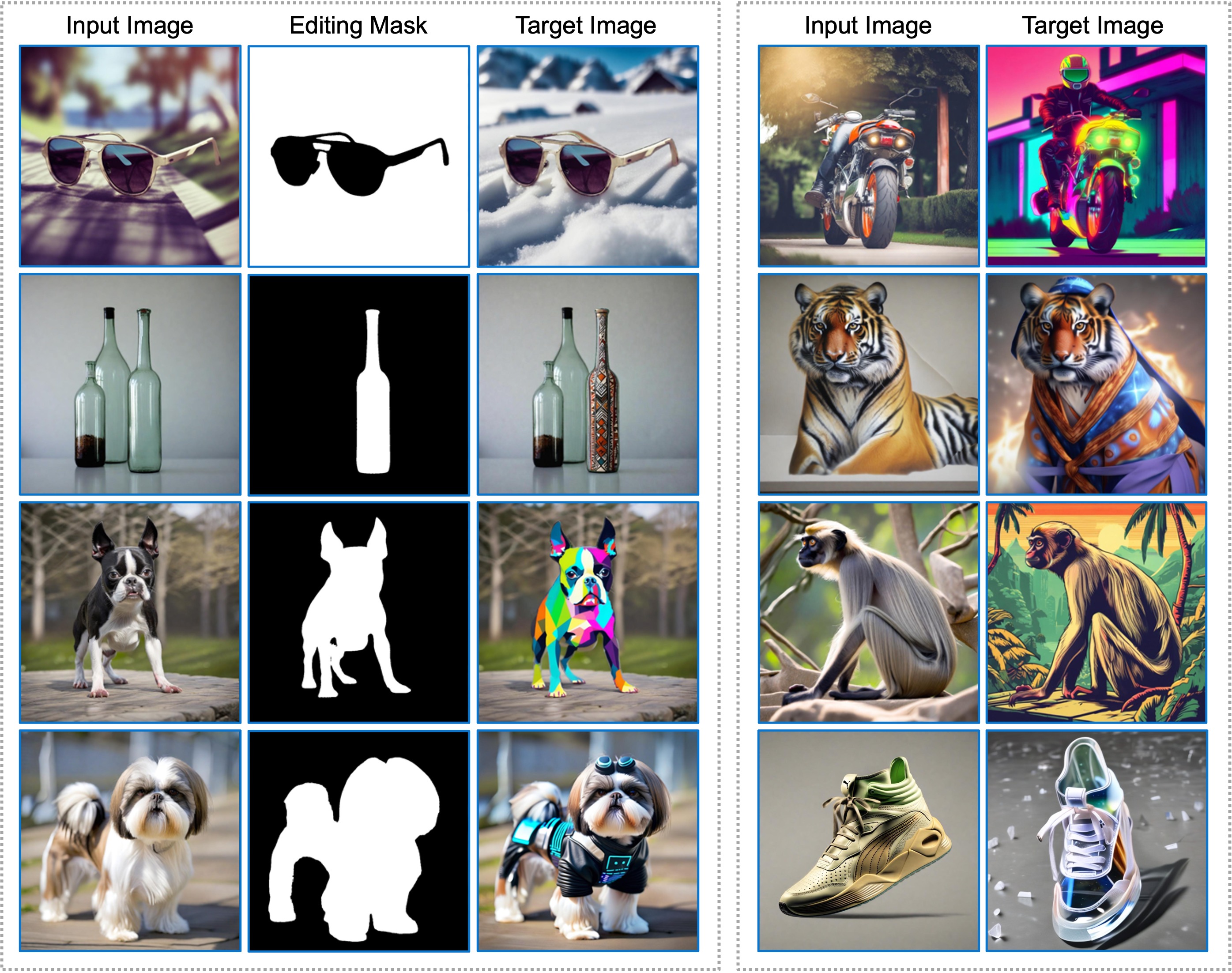}
    \vspace{-0.05in}
    \caption{Examples of the proposed dataset, including image editing (left) and generation (right) samples. Text prompts are not shown here due to the limited space. }
    \label{fig:data_example}
    \vspace{-0.1in}
\end{figure}

To obtain an image pair, we randomly sample a subject image and a text prompt, then generate the input and target image with the proposed framework. 
Furthermore, we also construct image editing pairs, where the input and target image only have local difference. Specifically, we use Grounded-SAM~\citep{ren2024grounded} to obtain subject masks, and combine the proposed framework with Blended Diffusion~\citep{avrahami2023blended, avrahami2022blended} to obtain target image with local changes. In editing pairs, we directly use the subject image as input image rather than generating another one with View Generator. 
The reason of constructing editing pairs is that we expect the resulting model trained on our final dataset is capable of both subject-driven image editing and generation, so that the users have better control over the generated images. Some data examples are provided in Figure \ref{fig:data_example}. 

After obtaining a large amount of samples, we apply automatic data filtering on generated pairs to further improve data quality. The generated pairs will first be filtered by the DINO similarity between input and target images to filter out pairs containing dissimilar subjects. 
The CLIP~\citep{radford2021learning} similarity between target image and text prompt will be used to filter out low-quality samples which are not text-aligned. In practice, we find that setting CLIP and DINO threshold to be 0.3 and 0.6 respectively normally leads to high-quality image pairs.

After filtering, we obtain a large-scale dataset Toffee-5M, comprising 4.8 million image pairs including 1.6 million image editing pairs with associated editing masks. The taxonomy is shown in Figure \ref{fig:dataset}, where
the image changes are categorized into the following categories: style change, background change, color change, texture change, element addition and removal.

\subsection{Unified Model for Subject-Driven Generation}
With the constructed dataset, we would like to obtain a model which is capable of both subject-driven image editing and generation. The model is expected to perform zero-shot editing and generation, without any test-time fine-tuning. Since our Toffee-5M dataset contains both image editing and generation pairs, we expect our model to be able to handle both cases within single network. Furthermore, input and target image from our generation pairs may have view and pose change. From the user's perspective, we also want to have the flexibility to control those changes during inference.

\begin{figure}[t!]
    \centering
    \includegraphics[width=1.0\linewidth]{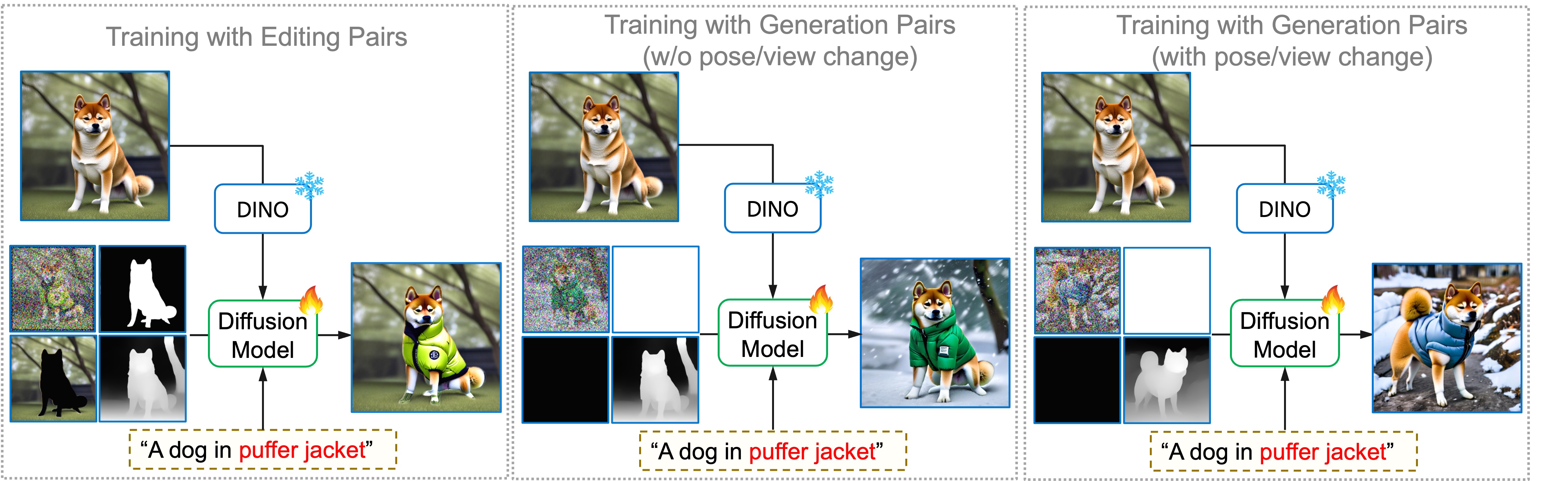}
    \vspace{-0.1in}
    \caption{Illustration of training single ToffeeNet model with both editing and generation pairs.
    }
    \label{fig:model}
    \vspace{-0.15in}
\end{figure}

We propose ToffeeNet, which is shown in the Figure \ref{fig:model}. We concatenate\footnote{In the case of Latent Diffusion Model like Stable Diffusion, the concatenation occurs in the latent space of the pre-trained Variational Auto-encoder.} editing mask, masked image and the noisy image of time t along channel dimension before feeding them into the diffusion model. The depth map of the target image is injected into the diffusion model via a ControlNet~\cite{zhang2023controlnet}. Specifically, for generation pairs, the editing mask is an all-white image, while the masked image is completely black. The DINO embedding of input image is introduced into the diffusion model via cross-attention layers. The corresponding cross-attention layer outputs of DINO and text embedding will be added in an element-wise manner, before being fed into next layer inside UNet.

\begin{figure}[t!]
    \centering
    \includegraphics[width=0.95\linewidth]{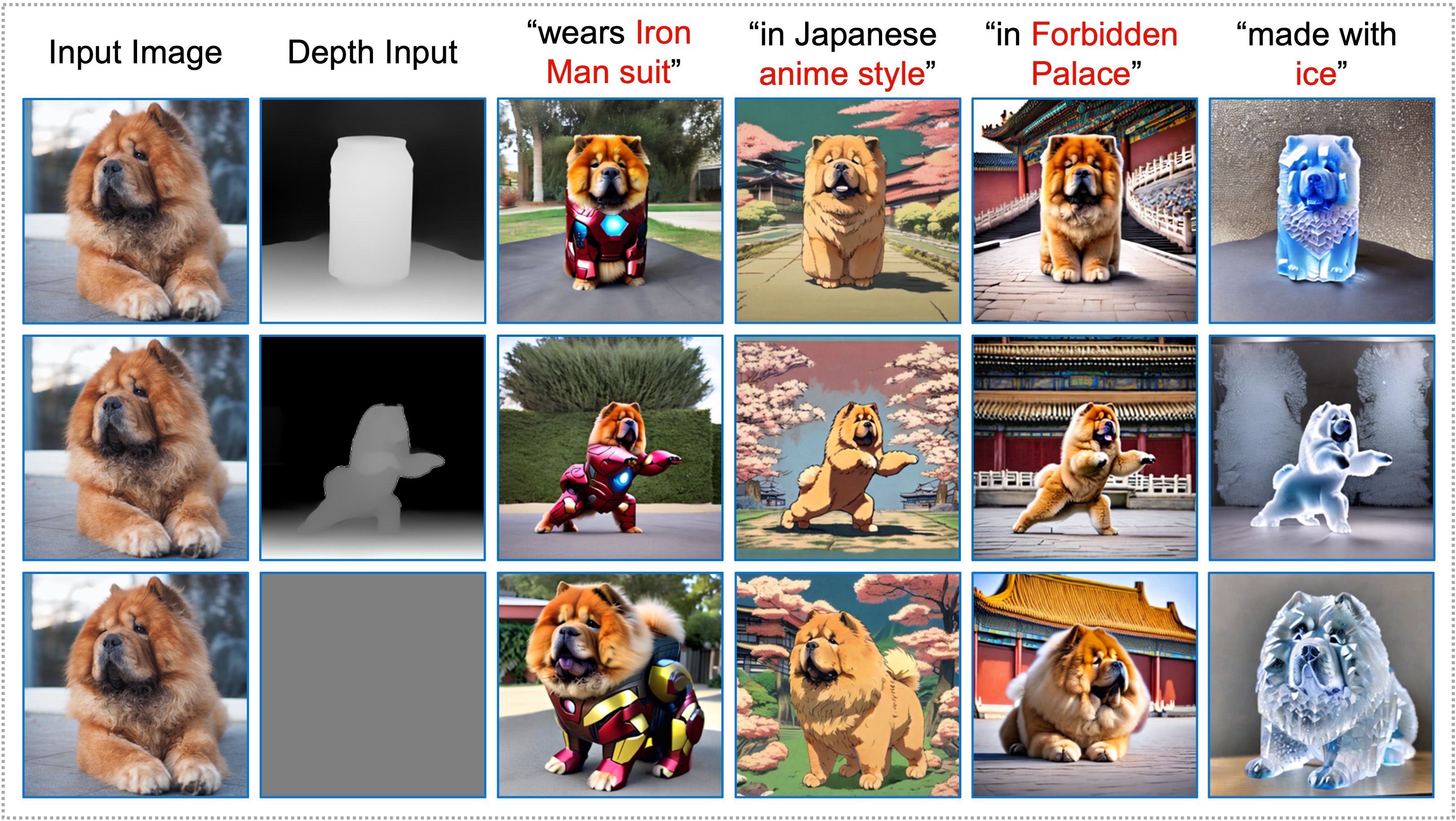}
    \vspace{-0.05in}
    \caption{We have the flexibility of controlling pose and view change by inputting a depth image.}
    \label{fig:depth_example}
    \vspace{-0.25in}
\end{figure}

During training, we replace the depth image by a constant image with a probability of 0.5. As a result, if we feed the constant image to the model during inference, the model will generate images with new views which are different from input image; if we feed the depth map of a given image (which can be the input image itself) into the model, the structural information will be preserved during generation. Some interesting examples are provided in Figure \ref{fig:depth_example} for better understanding, from which we can see that the generation follows the provided depth condition. 


\section{Experiment}
\subsection{Implementation Details}
We conduct all the experiments with PyTorch~\citep{paszke2019pytorch} on Nvidia A100 GPUs. AdamW~\citep{loshchilov2018decoupledAdamW} optimizer is used in all the model training. DINOv2-Giant~\citep{oquab2023dinov2}, which encodes an image as embedding $f(\rvx) \in \mathbb{R}^{257 \times 1536}$, is used in training Refiner, View Generator and ToffeeNet. DDIM sampling~\citep{song2020denoising} with 100 steps are used in evaluating all the models. We set the classifier-free guidance~\citep{ho2021cfg} to be 3.

Our Refiner is fine-tuned from a pre-trained Stable Diffusion XL~\citep{podell2023sdxl} on the union of CC3M dataset ~\cite{Sharma2018ConceptualCA} and generated subject images. The Refiner is trained for 200k steps, with a batch size of 64 and learning rate of 2e-5. 
Our View Generator is fine-tuned from a pre-trained Stable Diffusion 2~\citep{rombach2021high}, on MVImageNet dataset~\cite{yu2023mvimgnet}. The View Generator is trained for 200k steps, with a batch size of 128 and learning rate of 2e-5.

\begin{figure}[t!]
    \centering
    \includegraphics[width=0.99\linewidth]{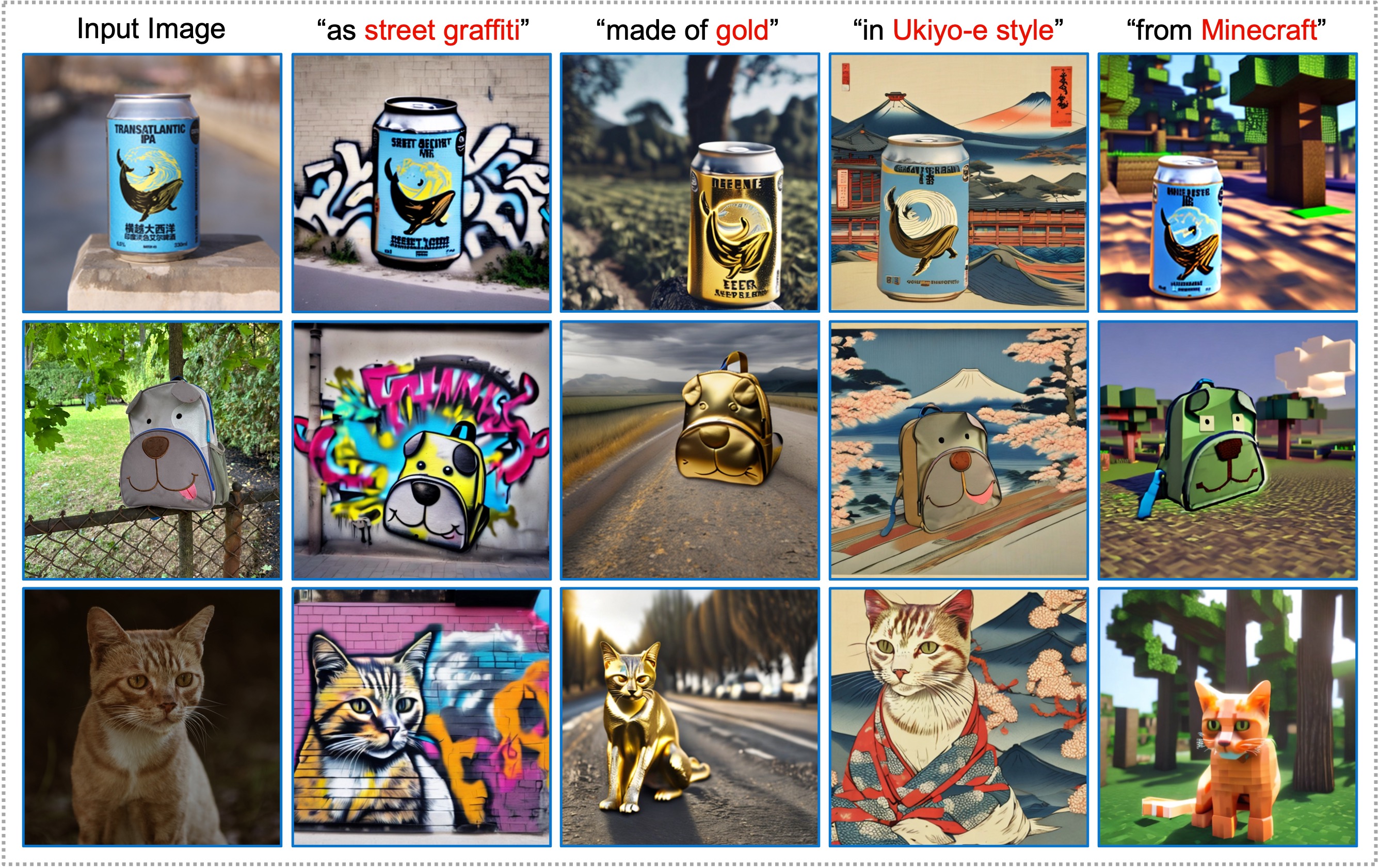}
    \vspace{-0.05in}
    \caption{Examples of image generation with ToffeeNet.}
    \label{fig:gen_example}
    \vspace{-0.1in}
\end{figure}

\begin{figure}[t!]
    \centering
    \includegraphics[width=0.95\linewidth]{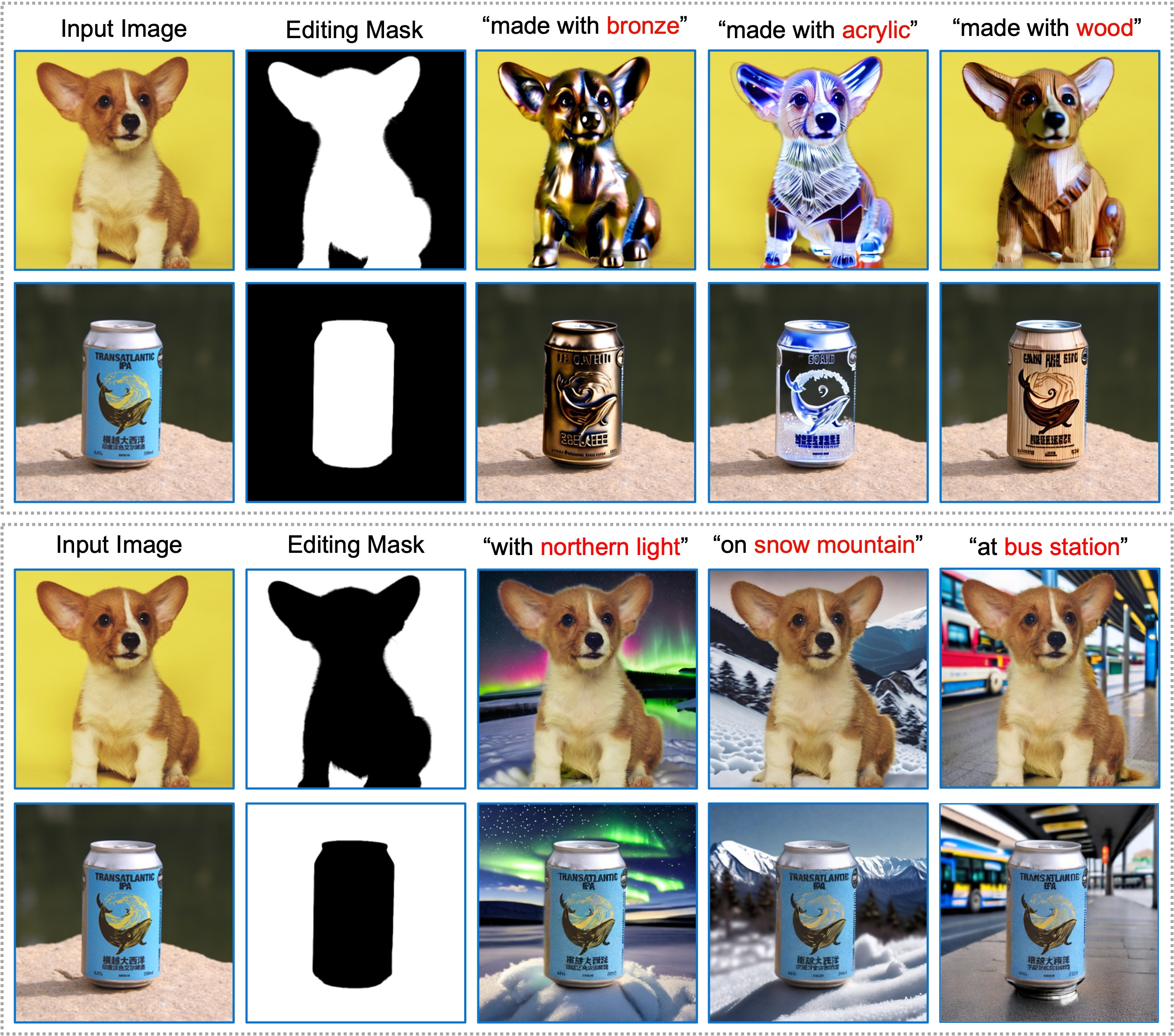}
    \vspace{-0.05in}
    \caption{Examples of image editing with ToffeeNet. We can control editing regions by feeding different masks into the model. The identity is well-preserved when the subject is being edited.}
    \label{fig:edit_example}
    \vspace{-0.1in}
\end{figure}

Our ToffeeNet is a fine-tuned Stable Diffusion 2, trained on the proposed Toffee-5M dataset for 100k steps. The batch size is set to be 128, learning rate is set to be 2e-5. Both DINO and text embeddings are independently dropped with a probability of 0.1 to enable classifier-free guidance~\citep{ho2021cfg}. After being trained on Toffee-5M dataset, the ToffeeNet can perform subject-driven image editing and generation in a tuning-free manner, which generates customized image with only 2 seconds given arbitrary subject image input.
Some image generation examples with our resulting model
is shown in Figure \ref{fig:gen_example}, some editing examples are provided in Figure \ref{fig:edit_example}. 

\subsection{Quantitative Results}
We conduct quantitative evaluation on DreamBench~\citep{ruiz2023dreambooth} following previous works. DreamBench contains 30 subjects and 25 text prompts for each subject. We select one input image for each subject and generate 4 images for each subject-prompt combination, resulting in 3,000 generated images. The generated images will be used to calculate metrics with pre-trained DINO ViT-S/16 and CLIP ViT-B/32. Specifically, image similarity is evaluated by average cosine similarity of image global embeddings between a generated image and all corresponding subject images. We use DINO and CLIP image encoder to extract these image embeddings, and denote corresponding scores as DINO and CLIP-I respectively. 
To evaluate whether the generation is text-aligned, we calculate the cosine similarity between embeddings of generated image and text prompt, which are extracted by pre-trained CLIP image and text encoder respectively. The image-text CLIP similarity is denoted as CLIP-T.

We compare our ToffeeNet with various methods including Textual Inversion~\citep{gal2022textualinversion}, DreamBooth~\citep{ruiz2023dreambooth}, CustomDiffusion~\citep{kumari2022multi}, BLIP-Diffusion~\citep{li2023blipdiffusion}, ELITE~\citep{wei2023elite}, Subject-Diffusion~\citep{ma2023subjectdiffusion}, Re-Imagen~\citep{chen2022reimagen}, SuTI~\citep{chen2023suti}, Kosmos-G~\citep{pan2023kosmos}, CAFE~\citep{zhou2023cafe}. The results are presented in Table \ref{tab:main_results}, where the results of corresponding methods are directly taken from their papers. For fair comparison, we also indicate whether a model is test-time tuning-free, and their diffusion model backbone. Note that although SuTI, CAFE can perform subject-driven generation without test-time fine-tuning, they require extra cost which is subject-level fine-tuning in dataset construction stage.

\begin{table}[t!]
  \caption{Quantitative results on DreamBench.}
  \label{tab:main_results}
  \centering
  \scalebox{0.95}{
  \begin{tabular}{lccccc}
    \toprule
    \multirow{2}{*}{Method} & \multirow{2}{*}{Backbone} & Test-time & \multirow{2}{*}{DINO $(\uparrow)$} & \multirow{2}{*}{CLIP-I $(\uparrow)$} & \multirow{2}{*}{CLIP-T$(\uparrow)$} \\
    & &Tuning-free & & & \\
    \midrule
    Real Images & - & - & 0.774 & 0.885 & - \\
    \midrule
    DreamBooth~\citep{ruiz2023dreambooth} & Imagen  & No & 0.696 & 0.812 & 0.306\\
    DreamBooth~\citep{ruiz2023dreambooth}  & Stable Diffusion & No & 0.668 & 0.803 & 0.305 \\
    Textual Inversion~\citep{gal2022textualinversion} & Stable Diffusion & No & 0.569 & 0.780 & 0.255 \\
    CustomDiffusion~\citep{kumari2022multi} & Stable Diffusion & No & 0.643 & 0.790 & 0.305 \\
    BLIP-Diffusion~\citep{li2023blipdiffusion} & Stable Diffusion & No & 0.670 & 0.805 & 0.302 \\
    \midrule
    Re-Imagen~\citep{chen2022reimagen} & Imagen & Yes & 0.600 & 0.740 & 0.270 \\
    SuTI~\citep{chen2023suti} & Imagen& Yes & 0.741 & 0.819 & 0.304 \\
    BLIP-Diffusion~\citep{li2023blipdiffusion} & Stable Diffusion& Yes & 0.594 & 0.779 & 0.300 \\
    ELITE~\citep{wei2023elite} & Stable Diffusion & Yes & 0.621 & 0.771 & 0.293 \\
    Subject-Diffusion~\citep{ma2023subjectdiffusion} & Stable Diffusion & Yes & 0.711 & 0.787 & 0.293 \\
    Kosmos-G~\citep{pan2023kosmos} & Stable Diffusion & Yes & 0.694 & 0.847 & 0.287 \\
    CAFE~\citep{zhou2023cafe} & Stable Diffusion & Yes & 0.715 & 0.827 & 0.294 \\
    ToffeeNet (Ours) & Stable Diffusion & Yes & 0.728 & 0.817 & 0.306 \\
    \bottomrule
  \end{tabular}
  }
  \vspace{-0.15in}
\end{table}

\begin{wrapfigure}{r}{0.5\textwidth}
    \vspace{-0.15in}
    \centering
    \includegraphics[width=0.99\linewidth]{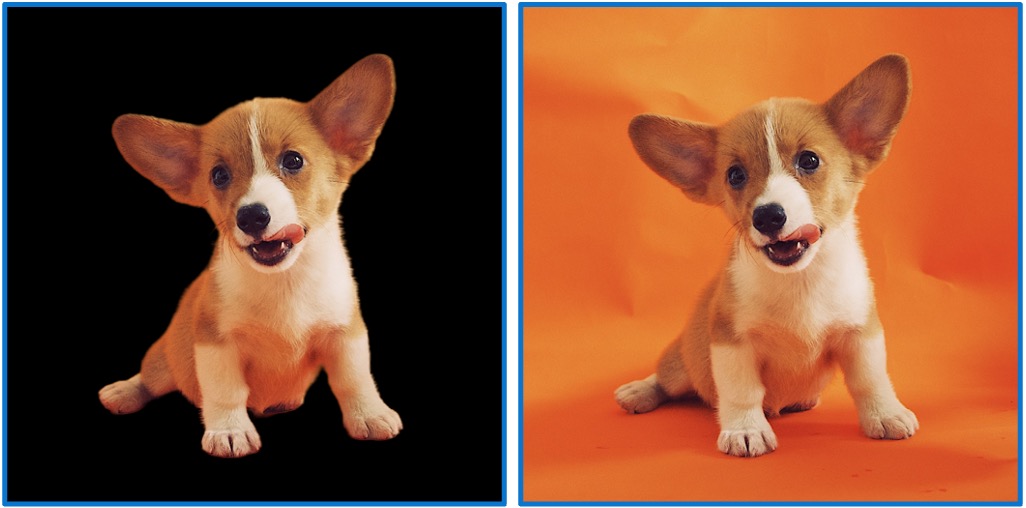}
    \caption{DINO and CLIP-I scores evaluated on this pair are 0.88 and 0.93, while the images contain the same subject without any change.}
    \label{fig:new_metric}
    \vspace{-0.15in}
\end{wrapfigure}
\subsection{Ablation Study}
\paragraph{New metrics} As discussed in previous works~\citep{chen2023disenbooth,ruiz2023dreambooth}, DINO and CLIP-I are flawed in evaluating subject similarity, because they can be influenced by background information. For example, images in Figure~\ref{fig:new_metric} contain the same dog, the left one is actually obtained from the right one using segmentation. However, the DINO and CLIP-I scores evaluated on this image pair are 0.88 and 0.93 respectively. Ideally, we expect the subject similarity to be 1 because these two images contains exactly the same subject. Meanwhile, the DINO and CLIP-T conflict with each other in the case of generation with background change, because a successful background change leads to high CLIP-T score but possibly low DINO score, even when the generation is perfect from human perspective.
Thus we expect better evaluation metrics. Specifically, we propose to use Seg-DINO and Seg-CLIP-I, which are evaluated by computing DINO and CLIP-I scores on the images obtained by applying segmentation on both subject and generated images. We use Grounded-SAM~\citep{ren2024grounded} for segmentation on both subject and generated images. Seg-DINO and Seg-CLIP-I will be applied in all ablation studies.

\paragraph{Model variants} Recall that our ToffeeNet is trained on all the samples from Toffee-5M dataset. We also train two variants, denoted as ToffeeNet-E and ToffeeNet-G, which are trained with only editing or generation pairs respectively. Comparison between these models are provided in Table \ref{tab:ablation_variants}. Note that ToffeeNet is capable of both generation and editing task, both results are reported. Additionally, we present results for both scenarios in generation task: without view change where the generation is conditioned on the depth map of the input image, and with view change where the depth map is a constant image.
We found that although unmasked region of the input image is well kept in editing task, the model may add extra subject in the background when it tries to perform background change, which leads to slightly worse Seg-DINO score than generation task in Table \ref{tab:ablation_variants}.

\begin{table}[t!]
  \caption{Results of different model variants. We can manage changes in terms of view and pose, by providing either a depth map of input image or a constant image to the model.}
  \label{tab:ablation_variants}
  \centering
  \scalebox{0.95}{
  \begin{tabular}{lcccccc}
    \toprule
    Model & Task & View Change & Seg-DINO $(\uparrow)$ & Seg-CLIP-I $(\uparrow)$ & CLIP-T $(\uparrow)$ \\
    \midrule
    Real Images & - & - & 0.854 & 0.927 & - \\
    \midrule
    \multirow{3}{*}{ToffeeNet} & Editing & No & 0.801 & 0.876 & 0.310 \\
    & Generation & No & 0.803 & 0.874 & 0.306 \\
    & Generation & Yes &  0.787 & 0.865 & 0.312\\
    \midrule
    ToffeeNet-E & Editing & No & 0.798 & 0.877 & 0.312 \\
    \midrule
    \multirow{2}{*}{ToffeeNet-G} & Generation & No & 0.806 & 0.876 & 0.299 \\
    & Generation & Yes & 0.805 & 0.875 & 0.304\\
    \bottomrule
  \end{tabular}
  }
\end{table}

\paragraph{Training with reconstruction task} We test ToffeeNet variants obtained by replacing input subject image by target image with a probability of $p$ during training, by which we basically force the model to perform image reconstruction with probability $p$. We report the results of ToffeeNet in Table \ref{tab:ablation_reconstruction}. With the introduced reconstruction task, we observe improvements in Seg-DINO and Seg-CLIP-I as expected, because the model can learn better subject details from reconstruction task. However, the CLIP-T score will decrease when we increase $p$ as the model focuses more and more on reconstruction task and has difficulty in generating text-aligned images.

\begin{table}[t!]
  \caption{Results of forcing ToffeeNet to perform reconstruction with probability $p$ during training.}
  \label{tab:ablation_reconstruction}
  \centering
  \scalebox{0.95}{
  \begin{tabular}{lcccccc}
    \toprule
    Task & View Change & $p$ & Seg-DINO $(\uparrow)$ & Seg-CLIP-I $(\uparrow)$ & CLIP-T $(\uparrow)$ \\
    \midrule
    \multirow{3}{*}{Generation} & \multirow{3}{*}{No} & 0 & 0.803 & 0.874  & 0.306  \\
     & & 0.25 & 0.806 & 0.879  & 0.288  \\
     & & 0.5 &  0.818 &  0.882 & 0.286 \\
     \midrule
     \multirow{3}{*}{Generation} & \multirow{3}{*}{Yes} & 0 & 0.787& 0.865  & 0.312  \\
     & & 0.25 & 0.805 & 0.876  & 0.301  \\
     & & 0.5 &  0.815 &  0.880 & 0.294 \\
     \midrule
    \multirow{3}{*}{Editing}& \multirow{3}{*}{No}& 0 & 0.801 & 0.876 & 0.310 \\
     & & 0.25 & 0.807 &  0.880 & 0.308 \\
     & & 0.5 &  0.812 &  0.881 & 0.300 \\
    \bottomrule
  \end{tabular}
  }
\end{table}

\paragraph{Subject image pre-processing} 
One intriguing question is whether we should use the entire image as input for the DINO encoder or if we should use only the segmented subject from the image. To address this question, we test two ToffeeNet variants trained using different inputs for the DINO encoder: one with the whole subject image and the other with the segmented subject. $p$ is set to be 0 for both models. We did not observe significant differences: in the generation task, the model trained with the whole image leads in Seg-DINO and Seg-CLIP-I by 0.003 and 0.001, respectively, while showing slightly worse performance in CLIP-T by only 0.001. 
In the editing task, both models achieve nearly the same performance.

\paragraph{Comparison with InstructPix2Pix} Because the proposed framework can be used to construct both image generation and editing pairs, we are interested in comparison with related editing method such as InstructPix2Pix~\citep{brooks2023instructpix2pix}, which is trained on synthetic dataset generated with Prompt-to-Prompt~\citep{hertz2022prompt}. The comparison is provided in Table \ref{tab:ablation_comparison_with_ip2p}. InstructPix2Pix proposes to use a classifier-free guidance with two conditional guidances, thus we report their results with different hyper-parameters for fair comparison. From the result we can see that InstructPix2Pix fails to maintain subject identity and obtain good text-alignment at the same time. Meanwhile, our proposed dataset construction is designed to preserve the subject identity, thus ToffeeNet can obtain text-aligned results without changing the identity too much.
Furthermore, the proposed Refiner can also be used to refine the training pairs in InstructPix2Pix, which means the proposed method can be seamlessly combined with others.
\begin{table}[t!]
  \caption{Comparison of the proposed method with InstructPix2Pix.}
  \label{tab:ablation_comparison_with_ip2p}
  \centering
  \scalebox{0.95}{
  \begin{tabular}{lcccccc}
    \toprule
    \multirow{2}{*}{Method} &Image & Text & \multirow{2}{*}{Seg-DINO $(\uparrow)$ } & \multirow{2}{*}{Seg-CLIP-I $(\uparrow)$ }  & \multirow{2}{*}{CLIP-T $(\uparrow)$ } \\
    & Guidance & Guidance & & & & \\
    \midrule
    \multirow{4}{*}{InstructPix2Pix} & 1.2 & 7.5 & 0.692 & 0.833 & 0.310 \\
    & 1.5 & 7.5 & 0.748 & 0.849 & 0.294 \\
     & 1.8 & 7.5 & 0.806 & 0.874 & 0.279 \\
    \midrule
    ToffeeNet & - & - & 0.801 & 0.876 & 0.310 \\
    ToffeeNet-E & - & - & 0.798 & 0.877 & 0.312 \\
    \bottomrule
  \end{tabular}
  }
\end{table}

\begin{figure}[t!]
    \centering
    \includegraphics[width=0.99\linewidth]{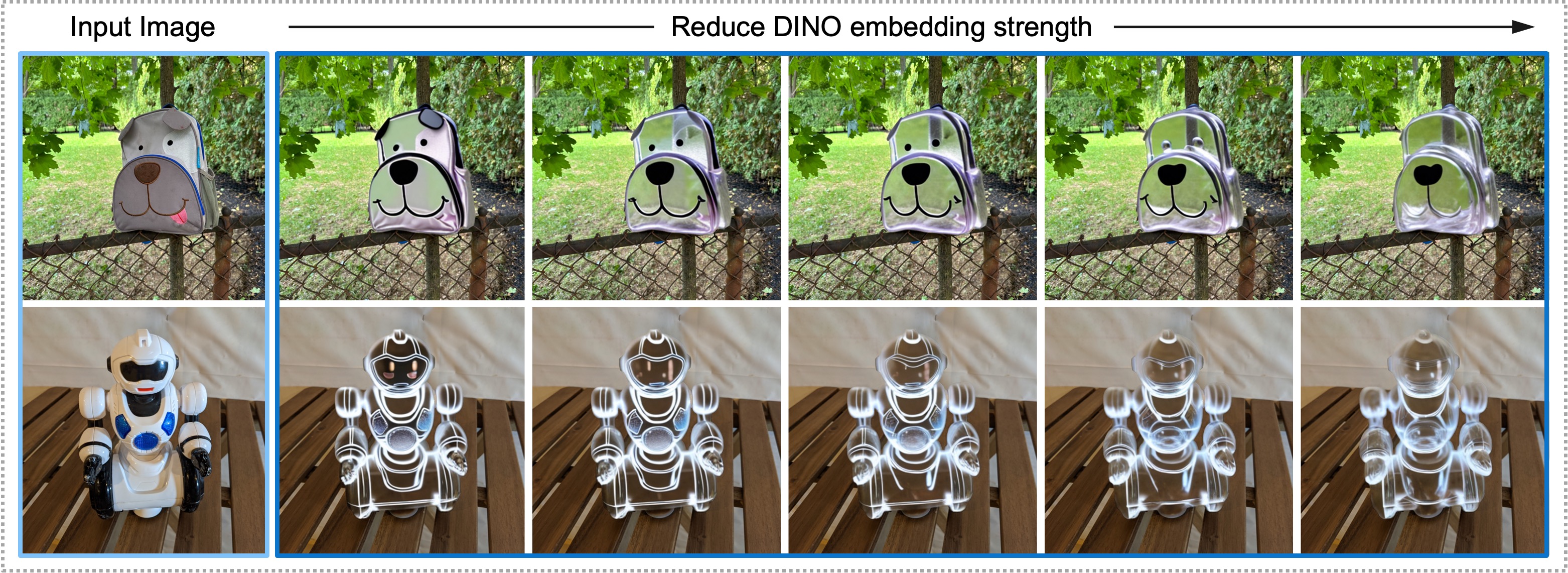}
    \vspace{-0.05in}
    \caption{Results with different DINO embedding strength, prompt used here is ``transparent object''.}
    \label{fig:ablation_dino_strength}
    \vspace{-0.2in}
\end{figure}

\paragraph{DINO embedding strength} During training, DINO embeddings of input image will be fed into cross-attention layers, whose outputs will be element-wisely added with outputs from cross-attention layers for text embeddings. At test-time, we can scale the DINO-related cross-attention outputs by $0 \leq \lambda \leq 1$. As $\lambda$ decreases, the generation will be less conditioned on input image and more conditioned on text prompt. Some examples are shown in Figure \ref{fig:ablation_dino_strength}, from which we can find that the objects become more transparent when we decrease the $\lambda$. However, some subject details will be lost when $\lambda$ becomes too small. In practice, we found $\lambda \in \left[0.7, 1.0\right]$ works well in most scenarios.

\section{Related Works}
There are numerous existing works in subject-driven text-to-image generation domain. Some methods require test-time optimization or fine-tuning. For instance, DreamBooth~\citep{ruiz2023dreambooth} and CustomDiffusion~\citep{kumari2022multi} propose to fine-tune pre-trained diffusion model on testing images. Textual Inversion~\citep{gal2022textualinversion} proposes to represent the subject with an embedding learned via optimization, which is then extended to multiple embeddings in \citep{voynov2023p+}.  

Aforementioned works are often time-consuming and require at least minutes before generating images for the subject. To tackle this challenge, some works try to train an image encoder~\citep{gal2023e4t, wei2023elite} so that the subject can be readily represented as embeddings at test-time. Instead of simply training an encoder for a frozen text-to-image model. Some works try to align pre-trained image encoders with diffusion models. For example, Subject-Diffusion~\citep{ma2023subjectdiffusion} introduces trainable adapter into diffusion model, and fine-tunes the text-to-image generation model while keeping image encoder frozen. Some works try to align language models with diffusion models: Kosmos-G~\citep{pan2023kosmos} introduces an AlignerNet on top of large language models (LLMs) to introduce multimodal information into pre-trained diffusion model; CAFE~\citep{zhou2023cafe} fine-tunes a LLM so that it can interact with users through conversation and predict semantic embeddings to guide the generation process of diffusion model.
There are also some works like Re-Imagen~\citep{chen2022reimagen} and SuTI~\citep{chen2023suti} which adopt a retrieval-augmented approach to condition the generation on retrieved images, so that the performance can be enhanced. 

Recent works like SuTI~\citep{chen2023suti} and CAFE~\citep{zhou2023cafe} have shown the importance of constructing high-quality synthetic dataset in subject-driven generation. With a high-quality dataset, impressive results are obtained in a test-time tuning-free manner, outperforms previous methods in terms of both efficiency and effectiveness. Inspired by SuTI and CAFE, we focus on improving the efficiency of constructing these synthetic dataset. Compared to existing methods, our proposed Toffee is a much more efficient method to obtain large-scale dataset for subject-driven image editing and generation.
\section{Limitation and Broader Impact}
Subject-driven image editing and generation have significant potential in real-world applications, as it can help users in generating creative images without expert knowledge. However, related methods can also lead to potential misinformation, abuse and bias. It is crucial to have proper supervision in constructing dataset, training model and applying these methods in real-world applications. In our work, all the training images are generated from pre-trained Stable Diffusion. By manually designing subject classes and filtering all the text prompts before generating Toffee-5M dataset, we try to avoid generating potential harmful and sensitive information.

One limitation of our proposed method is that the View Generator fails in certain cases. This is because our View Generator is trained on MVImageNet~\citep{yu2023mvimgnet}, which contains images across various object classes while has few samples for certain categories such as real world animals. As a result, the View Generator sometimes fails to generate new views for input animal image. We believe that the View Generator can be improved by training with a better multi-view image dataset.

\section{Conclusion}
We propose Toffee, a novel framework which can efficiently construct high-quality dataset for subject-driven image editing and generation tasks. Compared to previous methods which require O(N) fine-tuning steps to generate samples for a dataset with N subjects, our Toffee only needs O(1) fine-tuning steps. A large-scale dataset Toffee-5M is constructed, containing millions of image editing and generation pairs. We also propose a unified model named ToffeeNet, which is able to perform both image editing and generation. Training the ToffeeNet on our Toffee-5M dataset leads to competitive results for subject-driven text-to-image generation without any testing-time fine-tuning, illustrating the effectiveness of the proposed framework.
\clearpage

{
    \small
    \bibliographystyle{plain}
    \bibliography{main}
}

\end{document}